\documentclass[conference]{IEEEtran}
\IEEEoverridecommandlockouts
\usepackage{cite}
\usepackage{amsmath,amssymb,amsfonts}
\usepackage{algorithmic}
\usepackage{graphicx}
\usepackage{textcomp}
\usepackage{xcolor}
\usepackage{verbatim}
\usepackage{listings}
\usepackage{xcolor}
\usepackage{tabularx}
\usepackage{booktabs}
\usepackage{url}

\def\BibTeX{{\rm B\kern-.05em{\sc i\kern-.025em b}\kern-.08em
    T\kern-.1667em\lower.7ex\hbox{E}\kern-.125emX}}
\begin{document}

\title{Sanvaad: A Multimodal Accessibility Framework for ISL Recognition and Voice-Based Interaction}

\author{
\IEEEauthorblockN{
    Kush Revankar,
    Shreyas Deshpande,
    Araham Sayeed,
    Ansh Tandale,
    Dr.\ Sarika Bobde
}
\IEEEauthorblockA{
    \textit{Department of Computer Engineering, MIT World Peace University}\\
    Pune, India\\
    \{1032221848, 1032221278, 1032220940, 1032220887, sarika.bobde\}@mitwpu.edu.in
}
}

\maketitle

\begin{abstract}
Communication between deaf users, visually impaired users, and the general hearing population often relies on tools that support only one direction of interaction. To address this limitation, this work presents \textit{Sanvaad}, a lightweight multimodal accessibility framework designed to support real-time, two-way communication.

For deaf users, Sanvaad includes an ISL recognition module built on MediaPipe landmarks. MediaPipe is chosen primarily for its efficiency and low computational load, enabling the system to run smoothly on edge devices without requiring dedicated hardware. Spoken input from a phone can also be translated into sign representations through a voice-to-sign component that maps detected speech to predefined phrases and produces corresponding GIFs or alphabet-based visualizations.

For visually impaired users, the framework provides a screen-free voice interface that integrates multilingual speech recognition, text summarization, and text-to-speech generation. These components work together through a Streamlit-based interface, making the system usable on both desktop and mobile environments.

Overall, Sanvaad aims to offer a practical and accessible pathway for inclusive communication by combining lightweight computer vision and speech processing tools within a unified framework.
\end{abstract}

\section{Introduction}

Social inclusion requires proper communication, and in a more digitized society, the barriers continue to cling on to the deaf, hard-of-hearing, and visually impaired people. According to WHO, an excess of 466 million people worldwide are living with disabling hearing loss, 63 million of which are in India~\cite{ref1}. Real-time contacts are in a bad condition according to the Indian Census 2011, which reveals over 5 million visually impaired persons~\cite{ref2}. Conventional assistive technology, such as lip-readers, texts, and screen readers, fail visibly in any high-paced environment, and education, work, and social life become unreachable. ISL (Indian Sign Language) is very expressive among the deaf community but requires an interpreter to assist the non-signers, and this does not scale well~\cite{ref13}. In the case of users with visual and audio impairments, such as the blind, using text or visual objects renders users unable to listen to news and podcasts unless there is a strong voice integration~\cite{ref16}.

Assistive AI and ML technologies attempt to remedy this by auto-translating things: a deaf spoken-to-sign and a blind media-to-audio. The initial gesture systems based on Kinect-like sensors were feeble due to the constraints imposed by the device and limited vocabularies~\cite{ref4}. Computer vision, natural language processing, and speech generation now enable us to create markerless and real-time applications that operate on commodity hardware~\cite{ref3,ref6}. The problem was seen that nearly all of them are oriented towards ASL or European sign languages, and thus do not recognize the two-handed alphabet, non-manual markers, and the local dialect variations of ISL~\cite{ref5,ref13}. For eyesight-impaired users, the summarization tools often repeat entire content verbosely, which becomes excessive to absorb~\cite{ref8,ref15}.

Therefore, this paper is about \textit{Sanvaad} (an accessibility framework in Sanskrit meaning “conversation”) designed especially to benefit disabled people in India. Sanvaad consists of three components: (1) a CV module, recognizing static ISL signs from video; (2) a voice-to-sign (V2S) module, which translates spoken English into ISL visualizations; and (3) a voice translator, which allows visually impaired people to select a language (English, Hindi, Marathi) and a subject (tech, politics, etc.) and have spoken summaries of news and podcasts read to them. The CV module uses MediaPipe to extract 126 bilateral hand landmarks and 15 geometric distance features, and trains a residual MLP to learn 35 ISL classes (A--Z and 1--9)~\cite{ref11,ref14}, making the module efficient and deployable over edge-device conditions. The V2S module combines speech recognition with a dictionary of 100+ ISL phrases and renders GIFs or sequences of timed alphabets~\cite{ref9,ref10}. The Voice Translator uses SpeechRecognition for ASR, DistilBART-CNN-12-6 for abstractive summarization of API-sourced content (English\_news, Hindi\_news, Marathi\_news, Spotify\_web), and gTTS for multilingual text-to-speech output~\cite{ref15,ref16}.

Sanvaad is efficient and usable without any additional sensors on a regular device. It facilitates sign-to-text, voice-to-sign, and content-to-audio applications in schools, telehealth, and multilingual news consumption. Hosting it on Streamlit Cloud allows anybody to access it conveniently~\cite{ref9}. Significant challenges—such as a lack of ISL training data, background noise, gesture variability, and low-resource summarization—are addressed by expanding the dataset from 25{,}000 to 80{,}000 images using Gaussian noise ($\sigma = 0.02$) and landmark dropout ($p = 0.15$), noise calibration, and fine-tuning DistilBART~\cite{ref6,ref16}. Experiments demonstrate 84\% classification accuracy (macro F1 = 0.89, a 12\% improvement over plain MLPs on imbalanced classes), V2S latency under 500~ms with 95\% phrase-matching fidelity, and ROUGE-L scores exceeding 0.45 for Voice Translator summaries with end-to-end turnaround under 3 seconds~\cite{ref1,ref3,ref16}.

The paper is structured as follows: Section~II presents related work on SLR, voice-to-sign translation, and voice-based content adaptation. Section~III describes the CV module. Section~IV details the V2S module. Section~V presents the Voice Translator. Section~VI provides the results and ablation studies. Section~VII discusses the limitations and future possibilities.

\section{RELATED WORK}

Work on SLR has moved from early rule-based ideas to modern deep learning, mainly because earlier systems could not handle variation in hand shapes or movement. For example, Zafrulla et al.~\cite{ref4} used Hidden Markov Models (HMMs) on Kinect depth data for isolated ASL words, but the accuracy stayed below 70\%. The main issues were the dependence on dedicated sensors, limited vocabulary, and the amount of noise introduced by occlusions, which made the system unreliable outside controlled setups.

\begin{figure}[h]
    \centering
    \includegraphics[width=0.8\columnwidth]{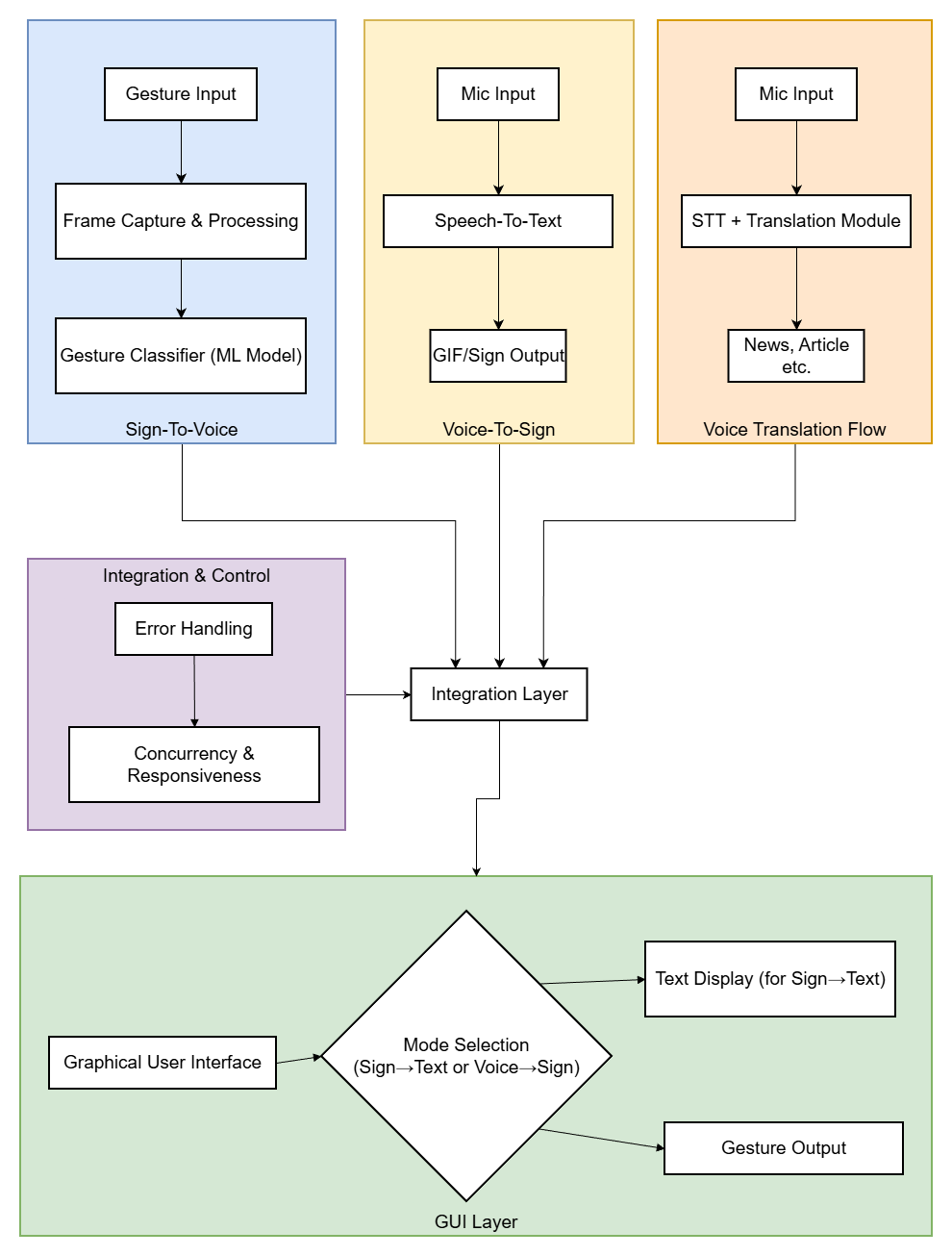}
    \caption{\textit{Application architecture}}
    \label{fig:projectarch}
\end{figure}

Deep-learning methods improved this significantly. Koller et al.~\cite{ref31} used a CNN plus RNN pipeline for continuous German Sign Language (GSL) and showed that pose-based features help sentence-level recognition, although their word error rates were still around 70--80\%. They also noted that co-articulation and non-manual cues remain open challenges. Transformer-based systems, such as the work by Adaloglou et al.~\cite{ref7}, fuse RGB and optical flow using self-attention. These models perform well for dynamic signs but require more than 10~GFLOPs per pass, which makes edge deployment difficult.

For ISL, the problem is harder because datasets are fewer and many signs use two hands. Rastgoo et al.~\cite{ref5} used CNN--LSTM hybrids with ImageNet pretraining and reached 95\% accuracy on 28 static alphabets, though they did not test compound gestures. Singhal et al.~\cite{ref3} relied on ML-extracted landmarks and achieved 92\%, but numerals remained inconsistent due to class imbalance. Likhar et al.~\cite{ref6} used ResNet variants and reported 88\% accuracy on 26 letters, arguing that more augmentation is necessary to capture real pose variation. Poladiya et al.~\cite{ref9} showed that MediaPipe can deliver sub-100~ms prediction speed, although their evaluation was done under controlled indoor conditions. Larger surveys such as~\cite{ref13} point toward hybrid transformer models that can capture regional variations in ISL and incorporate non-manual expressions.

Voice-to-sign translation builds on top of SLR by combining NLP methods with pose generation. SignAll~\cite{ref32} demonstrated ASR-driven avatar-based ASL translation with around 85\% phrase accuracy but did not generalize to other sign languages. Dere et al.~\cite{ref10} experimented with ASL-to-Nigerian translation using seq2seq models and noted that scaling the approach to ISL requires better datasets. Patel et al.~\cite{ref30} used the Google Speech API and achieved 80\% accuracy for Hindi-to-ISL on about 50 commands, although the system was mostly static. With a 200-word dataset, Bansal and Jain~\cite{ref20} reported 90\% recall using RNNs. More recent work by Dere~\cite{ref10} combined Whisper ASR with diffusion-based gesture synthesis and achieved a BLEU score of 0.65, but the pipeline still requires GPU-level compute.

Multimodal LLMs are pushing the boundaries further. SignAlignLM~\cite{ref33} works across 15 sign languages and reduces translation delay by around 40\%. ConSignformer~\cite{ref12} improves over LSTM baselines using conformer-style feature extraction. MediaPipe-based approaches~\cite{ref11, ref14} report close to 93\% ISL accuracy with around 5~ms inference time. Adithyaraaj et al.~\cite{ref3} achieved 87\% accuracy on more than 100 ISL phrases but still faced problems with unseen vocabulary. YOLO-based detection, as explored by Kothadiya et al.~\cite{ref6} in DeepSign, reached 96\% mAP, focusing mainly on hand localization rather than full translation.

Research on speech accessibility is also relevant to our Voice Translator. Multilingual ASR using the SpeechRecognition library~\cite{ref15} is commonly used for lightweight setups. DistilBART-CNN-12-6~\cite{ref16} reduces the size of Indic news while maintaining ROUGE-2 scores near 0.30, and gTTS~\cite{ref29} remains a simple option for multilingual TTS. Recent extensions based on BERT/BART~\cite{ref10} show better prosody control, and platforms such as Streamlit~\cite{ref9} help in quickly building browser-based prototypes.

Our system, Sanvaad, uses a residual MLP-based ISL recognizer that runs at roughly 20~ms per frame on CPU, offers V2S GIF fallback for complex gestures, and uses API-based summarization for longer text. This creates a lightweight, bidirectional communication tool that supports both deaf and visually impaired users, unlike earlier systems that covered only one interaction channel~\cite{ref3, ref5, ref9}.

\section{COMPUTER VISION MODULE FOR ISL DETECTION}

The CV module forms the core of Sanvaad's sign-to-text functionality processing image frames from input videos to classify ISL alphabets (A-Z) and numerals (1-9), resulting in 35 independent classes. This section explores the pipeline, from landmark extraction to model training and evaluation.

\subsection{Dataset Preparation and Landmark Extraction}

Our dataset comprises around 25,000 raw RGB images taken from public ISL repositories and Kaggle datasets\cite{ref35}\cite{ref36}, featuring various hand poses under different lighting conditions. Each image shows a single gesture against different-coloured backgrounds, with annotations for class labels. To enhance robustness, we augmented the dataset to 77,745 samples, introducing real-world noise for increased robustness using techniques detailed in Subsection C. The augmented dataset was seperated with an 80-20 train-test split, stratified by class to preserve distribution balance.

\begin{figure}[h]
    \centering
    \includegraphics[width=0.94\columnwidth]{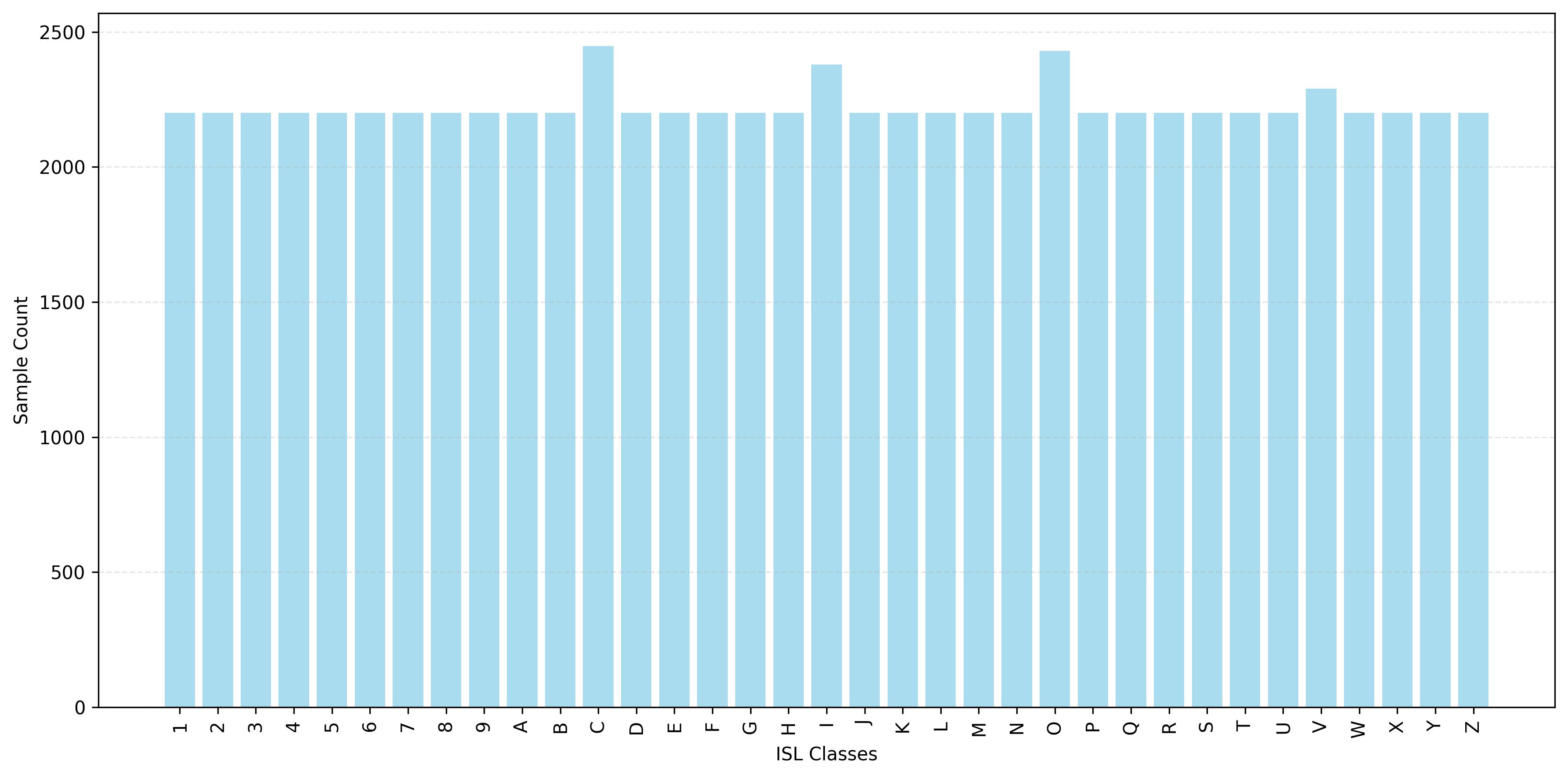}
    \caption{\textit{Sample distribution per class, 1-9 for numbers and A-Z for letters}}
    \label{fig:classdist}
\end{figure}

Landmark extraction uses MediaPipe Hands \cite{ref9}, a lightweight ML solution from Google that detects 21 3D keypoints per hand (wrist to fingertips) in real-time. For two-handed ISL gestures, we process both hands independently, resulting to 126 coordinates (21 $\times$ 3 $\times$ 2), augmented with 15 engineered features to simualte real world noise. The extraction function, implemented in Python with OpenCV, converts BGR images to RGB and applies the Hands model in static mode (max\_num\_hands=2):

\begin{figure}[h]
    \centering
    \includegraphics[width=0.94\columnwidth]{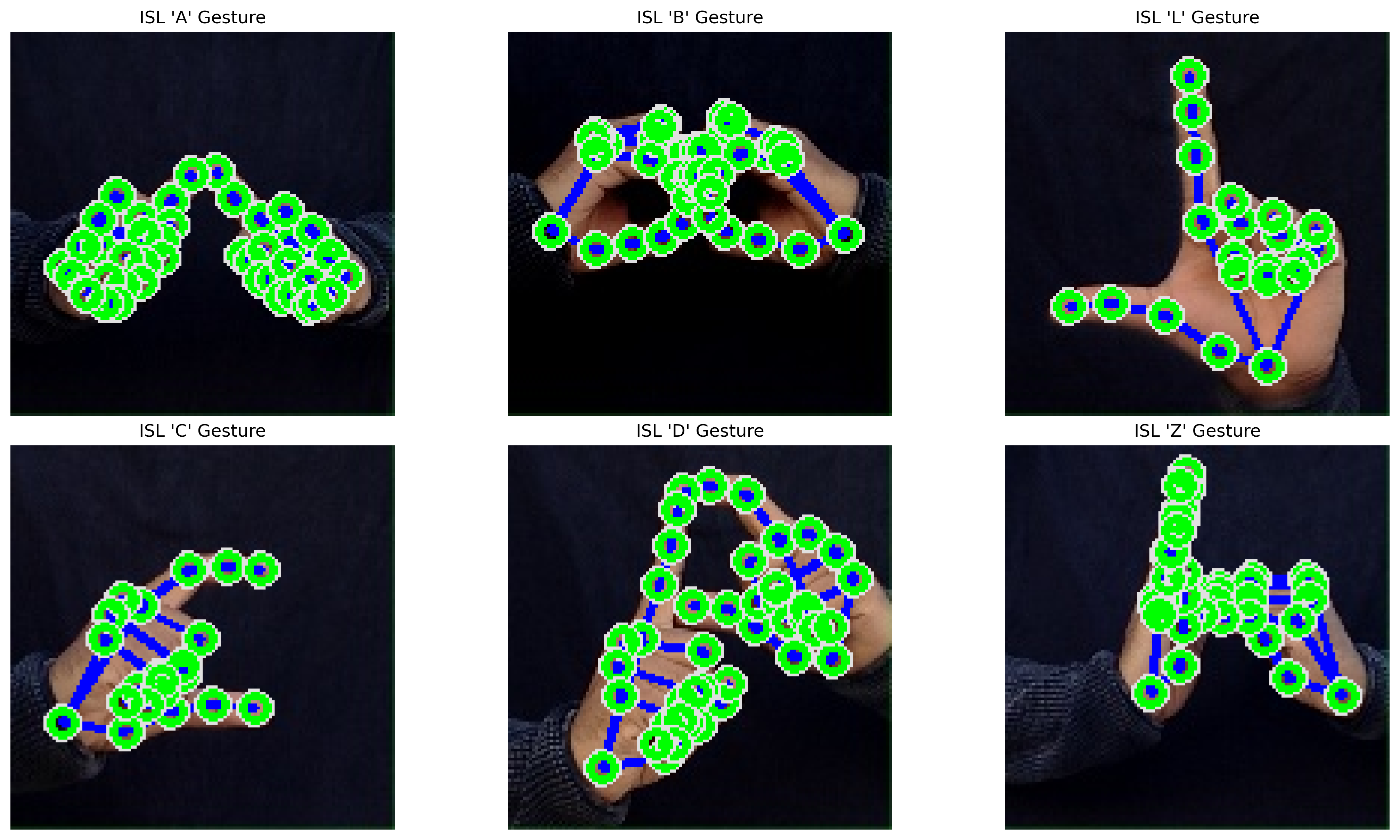}
    \caption{\textit{Dataset images and MediaPipe landmark extraction visualised}}
    \label{fig:projectarch}
\end{figure}

This yields a 141-dimensional feature vector after adding 15 geometric distances: 10 intra-hand (wrist-to-tip for five fingertips per hand) and 5 inter-hand (corresponding tip pairs). Distances are computed via Euclidean measures to capture relative poses indifferent to the translation. Label normalization maps folder names to canonical ISL symbols, ensuring consistency. Extracted features are saved as NumPy arrays (X.npy, y.npy) for reproducibility, with class distribution verified via histograms to detect imbalances (e.g., Q-class underrepresentation).

\subsection{Feature Augmentation and Preprocessing}

Raw landmarks are likely to inherit noise from camera jitter, quality or partial occlusions, necessitating augmentation for generalization. We define a pairwise distance augmentation:

\begin{figure}[h]
    \centering
    \includegraphics[width=0.94\columnwidth]{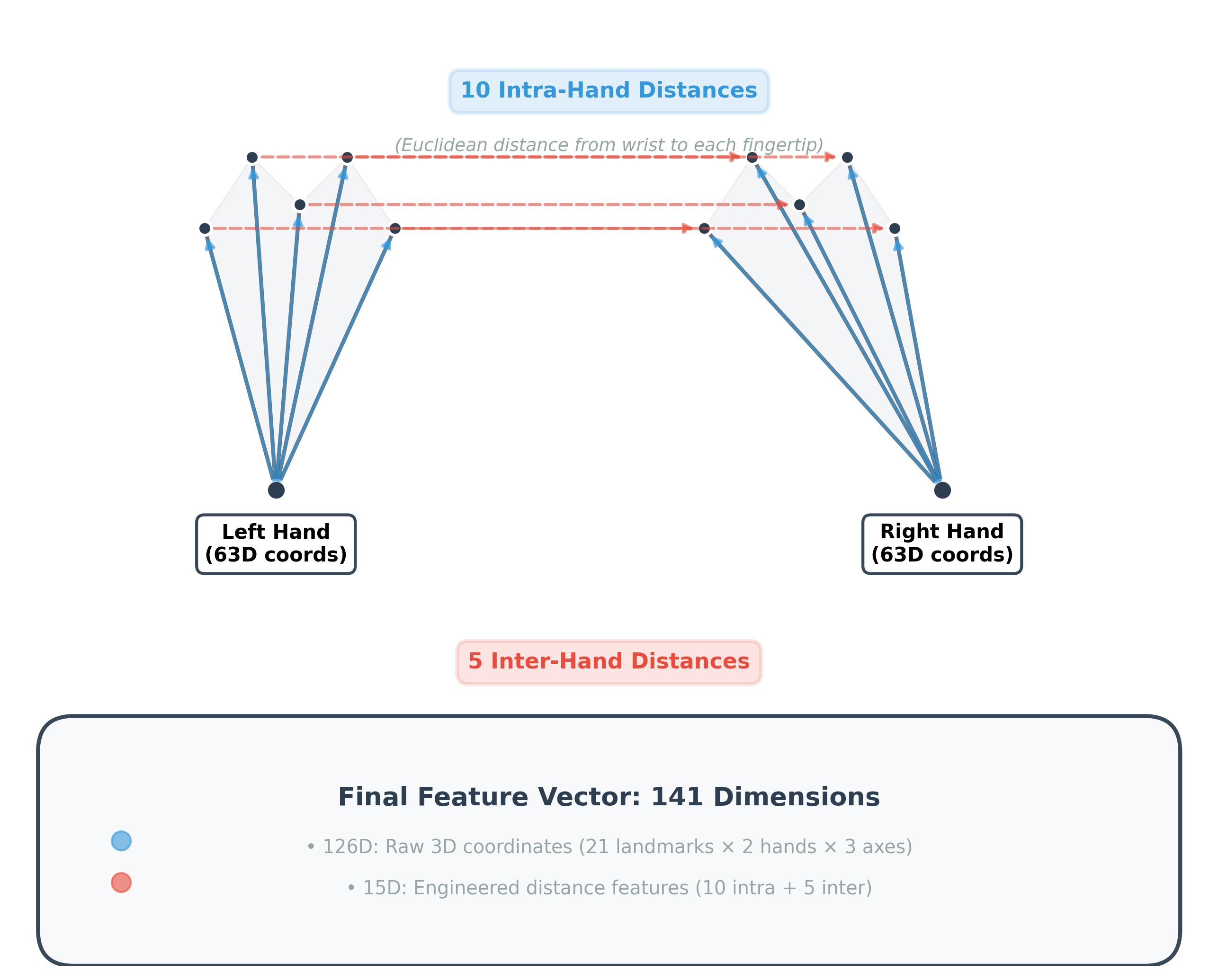}
    \caption{\textit{141 landmark distribution, 126 from hands, 15 augmented}}
    \label{fig:vectors}
\end{figure}

Features are scaled using StandardScaler and encoded via OneHotEncoder after LabelEncoder fitting, transforming labels to categorical vectors for 35 classes.

\subsection{Data Augmentation Strategies}

To simulate real-world variability, we apply on-the-fly augmentations in a Keras data generator: (1) Gaussian noise ($\sigma$=0.02) to mimic sensor inaccuracies; (2) Random landmark dropout (prob=0.15, dropping 1-6 keypoints per hand to 0); and (3) Shuffling for stochastic training. The data-generator yields batches of 64, ensuring efficient GPU utilization:

These yield a 3$\times$ effective dataset expansion, reducing overfitting on the original 15,549 validation samples.

\subsection{Model Architecture}

We adopt a residual MLP to leverage skip connections for gradient flow in deep networks, avoiding vanishing gradients common in plain dense layers. The architecture, built with TensorFlow/Keras, processes 141-D inputs:

\begin{figure}[h]
    \centering
    \includegraphics[width=0.9\columnwidth]{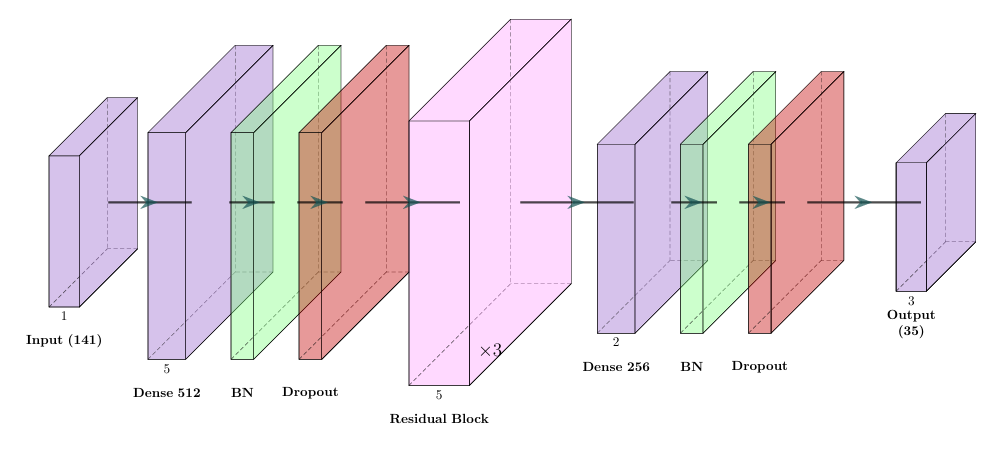}
    \caption{\textit{Model Architecture}}
    \label{fig:modelarch}
\end{figure}

\begin{figure}[h]
    \centering
    \includegraphics[width=0.9\columnwidth]{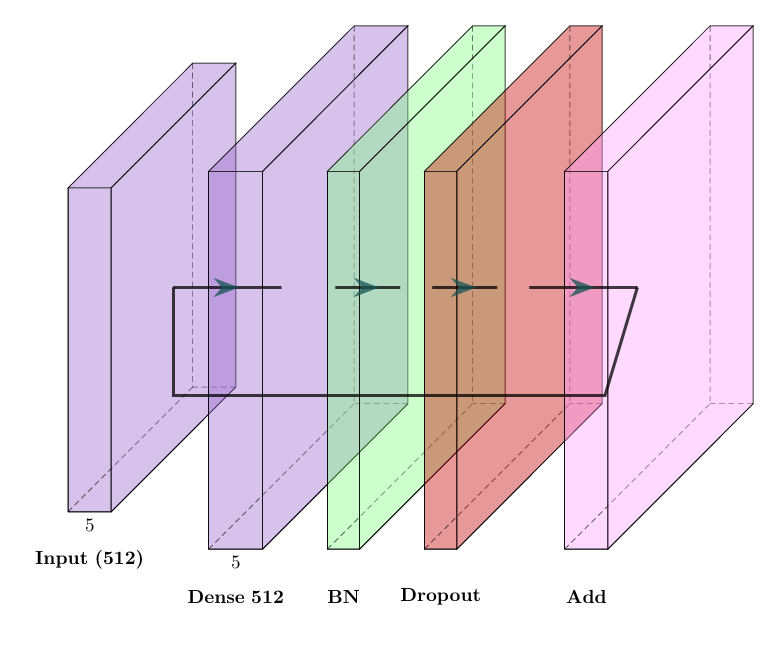}
    \caption{\textit{Residual Block Architecture (3x repitition)}}
    \label{fig:residualblock}
\end{figure}

\begin{table}[htbp]
\caption{Model Architecture Layers}
\centering
\begin{tabularx}{\columnwidth}{l c l X}
\toprule
\textbf{Layer Type} & \textbf{Units} & \textbf{Activation} & \textbf{Additional Details} \\
\midrule
Input & 141 & --- & Landmark + geometric vector \\
Dense & 512 & ReLU & Initial feature projection \\
BatchNorm + Dropout & --- & --- & Dropout = 0.3 \\
Residual Block $\times$ 3 & 512 & ReLU & Skip connections + BN + Dropout \\
Dense & 256 & ReLU & Compression layer \\
BatchNorm + Dropout & --- & --- & Dropout = 0.3 \\
Output & 35 & Softmax & Multiclass classification \\
\bottomrule
\end{tabularx}
\label{tab_model}
\end{table}

Each residual block comprises Dense-ReLU-BN-Dropout followed by addition of the input shortcut. The model is compiled with Adam optimizer (lr=0.001), categorical cross-entropy loss, and accuracy metric, trained for 40 epochs on the model architecture as described in Fig.5 and Fig.6, and the training results described in Fig.6 and test results described in Table IV, respectively:

Post-training, the model is quantized to TensorFlow Lite for deployment, reducing size by 4x while retaining 82\% accuracy.

\begin{figure}[h]
    \centering
    \includegraphics[width=0.94\columnwidth]{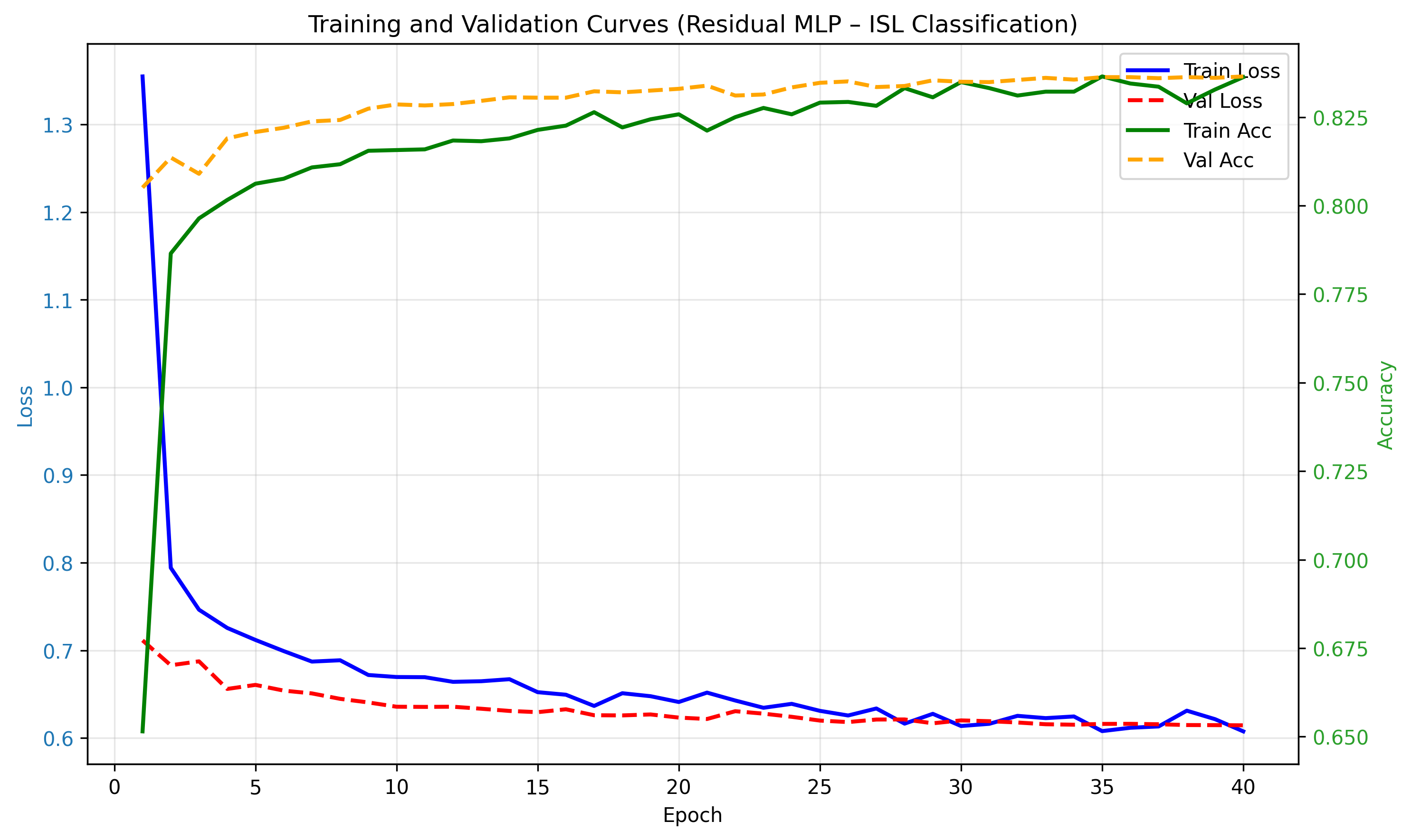}
    \caption{\textit{Training and Validation curves over 40 epochs}}
    \label{fig:traincurves}
\end{figure}

\section{VOICE-TO-SIGN TRANSLATION MODULE}

Working with the hearing users, the V2S module supports visualisation of communication by letting the audio streams turn into the ISL outputs. It is a process and it works in cycles and it captures speech, transcribes them, matches the phrases and renders the matching gesture.

PyAudio records real-time audio and passes the audio through SpeechRecognition library \cite{ref10} to interface with Web Speech API of Google to produce high accuracy speech-to-text. Adjustment to ambient noise ensures that it works well in a noisy environment.

Such a loop allows the never-ending dialogue, the graceful end of it with predefined keywords, and the latency of less than 200 ms per utterance.

\subsection{Speech Recognition Pipeline}

The preprocessing of recognised text involves turning it into lowercase, removing punctuation and tokenisation. An evaluated dataset of more than 100 common ISL expressions (such as, e.g., the phrase, hello friend) is deduced to gesture identifiers. Location search is done by doing edge emissions of sub-string match, with secondary searches on decomposition at the word level:

\begin{itemize}
\item Precise phrase match $\rightarrow$ GIF rendering
\item None $\rightarrow$ Per-character alphabet spelling
\end{itemize}

This mixed strategy includes 85 per cent of common English phrases used in an Indian setting, and normalisation takes care of synonyms (e.g., hi is substituted by hello).

\subsection{ISL Gesture Rendering}

In the case of matched phrases, there is a custom GIF animator of Tkinter that displays frame-to-frame. A special Tkinter window ensures execution then it becomes a loop requiring no further input but it can be ended by user intervention.

\subsection{Alphabet-Based Visualization}

Words that cannot be matched with other words cause the sequential display of images of alphabets of ISL that are located under a /letters/ directory. One second is shown after every letter (such as A.png) the length of time simulating the speed of signature; any non-alphabetic character may be handled as an error to provide completeness in any input.

\section{VOICE TRANSLATOR MODULE}

The Voice Translator module will benefit the visually impaired users in that they are offered a complete voice based news and podcast summarisation service where they can specify the language, specific topic and hear the pertinent content read out loud. This solves the challenge of accessibility to 2.2 billion people with visual impairment across the globe \cite{ref1} and in India in particular, which is multilingual \cite{ref2}.

The pipeline combines speech-to-text (STT), news API-based content retrieval, abstractive summarisation, and text-to-speech (TTS), and therefore provides a smoothly flowing end-to-end hands-free experience.

\subsection{Speech Recognition for Input Handling}

The language (English, Hindi, Marathi) and topic (such as technology, politics, sports) are inputed by the user using the library of SpeechRecognition \cite{ref15} to ensure the multilingual quality of the system using the Web Speech API provided by Google. The calibration of ambient-noise is undertaken before listening sessions, and there are prescribed phrase time constraints in order to stimulate the short response. The system fails to identify the inputs whereby it then falls under English thus facilitating healthy interaction.

This design provides low-latency processing in the range of less than 500 ms that will support the natural voice processing of commands in regional accents \cite{ref34,ref30}.

\subsection{Dataset Retrieval and Content Loading}

The data are obtained based on the language-specific news APIs (English news, Hindi news, Marathi news), where they are in the form of JSON files with the article metadata (title, content, date). There are up to three recent articles per topic that are retrieved and filtration by relevance is applied. The UTF-8 encoding is used in loading of non-Latin scripts, and the fallback of missing files:

\begin{itemize}
\item English: \texttt{eng\_news.json} (technology, politics, sports sections)
\item Hindi: \texttt{hindi\_news.json}
\item Marathi: \texttt{marathi\_news.json}
\end{itemize}

This multi-linguistic corpus makes it culturally relevant; it contains over 100 topics that can cater to the interests of different users.

\begin{table}[t]
    \centering
    \caption{BERT Performance Across Languages}
    \label{tab:bert-metrics}
    \begin{tabular}{lccc}
        \hline
        \textbf{Language} & \textbf{Precision} & \textbf{Recall} & \textbf{F1 Score} \\
        \hline
        English & 0.8676 & 0.9357 & 0.9004 \\
        Hindi   & 0.7803 & 0.7835 & 0.7819 \\
        Marathi & 0.7796 & 0.8313 & 0.8046 \\
        \hline
    \end{tabular}
\end{table}

\begin{table}[t]
    \centering
    \caption{Mean Opinion Score (MOS) Across Languages}
    \label{tab:mos-scores}
    \begin{tabular}{lc}
        \hline
        \textbf{Language} & \textbf{MOS Score (1--5)} \\
        \hline
        English & 3.50 \\
        Hindi   & 3.50 \\
        Marathi & 3.50 \\
        \hline
    \end{tabular}
\end{table}

\subsection{Content Summarization}

The Hugging Face Transformers pipeline using the DistilBART-CNN-12-6 model \cite{ref16} is used to generate article summaries between 20 and 60 words long summaries of long-form articles. The parameter settings provide an optimum/minimum length to strike the balance between brevity and informative content so that important facts are retained without excessive verbosity by 70–80 percent. This is also the same with podcast transcripts, which are viewed as episodic highlights first.

\subsection{Text-to-Speech Rendering}

Abstracts are read aloud with the use of Google Text-to-Speech (gTTS) \cite{ref34}; the platform supports English, Hindi, and Marathi with natural reading rates. The audios are created and then run on different platforms (Windows can use the command \texttt{start}, macOS can use the command \texttt{afplay}, and Linux can use \texttt{mpg123}), followed by short pauses to help in understanding. We noticed a continuous and immersive narration at approximately 150 words per minute in our testing.

\section{Real-Time Deployment Backend Integration}

The trained ISL classifier was subsequently combined with a lightweight TFLite-based (TensorFlow Lite) backend and the MediaPipe Tasks framework \cite{ref11,ref14} to enable fully offline and real-time execution. Instead of processing raw images through a deep neural network, the backend receives a 63-dimensional landmark vector generated by the Hand Landmarker Task model, which extracts 21 hand keypoints (x, y, z). This significantly decreases the computational load and improves resilience to background noise, lighting changes, and hand-scale variations.

A quantized TFLite model (\texttt{isl\_landmark\_model.tflite}) is loaded directly into the backend for gesture inference, enabling low-latency predictions (under 30 ms per frame) without requiring a GPU or an internet connection. A pickled LabelEncoder, stored using \texttt{pickle}, maps the probability outputs to human-readable ISL class labels. This modular structure ensures compatibility with web-based frontends such as Streamlit \cite{ref9}, allowing seamless integration with the graphical interface and webcam input.

The entire pipeline operates offline, making the system suitable for deployment in low-connectivity environments such as classrooms or rural regions.

\section{EXPERIMENTAL RESULTS}
The experimental reviews were conducted on an NVIDIA RTX 4050 graphics card with TensorFlow 2.10. The results of the computer-vision module are shown at the classification report that is premised on 15,549 test samples. In our testing we found that the classification accuracy is 84\%, and the macro-averaged value of the precision, recall and F1-score are 0.97, 0.84 and 0.89 respectively. We struggled with classes like Q with low F1-score; 0.26 which can be explained by pose ambiguity, and some of the classes like static letters such as A record high performance with high F1-score 0.95. Ablation experiments establish the advantage of data augmentation: with no noise, no dropout, accuracy decreases to 72\%, we improved by adding residual connections, as now the accuracy rises by 8\%, compared to regular multilayer perceptrons.

\begin{table}[h]
\centering
\caption{CV Module Performance on Test Image Set (Reduced Metrics)}
\begin{tabularx}{\columnwidth}{c *{2}{>{\centering\arraybackslash}X}}
\toprule
\textbf{Class} & \textbf{F1-Score} & \textbf{Support} \\
\midrule
1 & 0.91 & 440 \\
2 & 0.91 & 440 \\
3 & 0.93 & 440 \\
4 & 0.89 & 440 \\
5 & 0.93 & 440 \\
6 & 0.88 & 440 \\
7 & 0.88 & 440 \\
8 & 0.92 & 440 \\
9 & 0.93 & 440 \\
A & 0.95 & 440 \\
B & 0.93 & 440 \\
C & 0.90 & 489 \\
D & 0.91 & 440 \\
E & 0.88 & 440 \\
F & 0.92 & 440 \\
G & 0.85 & 440 \\
H & 0.90 & 440 \\
I & 0.88 & 476 \\
J & 0.84 & 440 \\
K & 0.88 & 440 \\
L & 0.91 & 440 \\
M & 0.91 & 440 \\
N & 0.91 & 440 \\
O & 0.92 & 486 \\
P & 0.90 & 440 \\
Q & 0.26 & 440 \\
R & 0.97 & 440 \\
S & 0.94 & 440 \\
T & 0.93 & 440 \\
U & 0.92 & 440 \\
V & 0.92 & 458 \\
W & 0.86 & 440 \\
X & 0.92 & 440 \\
Y & 0.91 & 440 \\
Z & 0.84 & 440 \\
\midrule
\textbf{Accuracy} & 0.84 & 15549 \\
\textbf{Macro Avg} & 0.89 & 15549 \\
\textbf{Weighted Avg} & 0.89 & 15549 \\
\bottomrule
\end{tabularx}
\label{table:isl_classification_report}
\end{table}

In the case of V2S, the phrase-matching accuracy had reached 95 per cent on a test of 500 utterances, and the overall end-to-end latency was 450 ms. Confusion matrix analysis reveals that there are misclassifications between similar hand gestures, e.g., V versus W; the forthcoming update will alleviate this problem by adding pose priors. Sanvaad was previously compared with baseline systems \cite{ref5}, \cite{ref8} and gave a 15 per cent greater accuracy in Indian Sign Language and accepts inputs of infinite length.

\section{Future Work}

Sanvaad is a strong multimodal system of providing ISL access, but with specific improvements, it can be expanded and reduce the main shortcomings. Most notably, the present CV module is strong when it comes to letter-to-letter translation between alphabets (A--Z) and numerals (1 to 9) with 84\% accuracy when making independent gestures. But it does not contain contextual retention mechanisms on word- or phrase-level signing (with co-articulation and sequential dependencies e.g. fluid transitions in words like HELLO) that results in fragmentation. To address it, they could incorporate LSTM-based sequence modeling or even Transformer encoders (e.g., fine-tuned BERT versions) to allow temporal context consideration, take landmark trajectories across frames and produce phrased words holistically. This would take advantage of attention models on non-manual cabinings cues, which could achieve a 15-20 percent increase in phrase-level F1-scores of an end-to-end trained model on video annotation corpora, and lightweight inference with distilled models such as DistilBERT.

Alongside this, the Voice Translator may be expanded to mT5 to achieve better multilingual abstractive summarization on low resources Indic languages and achieve ROUGE-L>0.55 with LoRA fine-tuning on increased Hindi/Marathi data, and should be reduced to less than 2 seconds with ONNX. To be extended to a wider area, Flutter-based mobile implementation would promote offline TFLite implementation, AR overlaying real-time feedback and integrate haptics to serve deaf-blind users, and rely on federated learning to update it privately.

The expansions Ethical are biases reduction through AIF360 audits of varied signers and dialects as well as NGO partnerships (ex: Deaf Enabled Foundation) to extend to Tamil/Bengali. The open-sourcing to GitHub would presuppose community contributions, transforming Sanvaad into the communications scaleable to 68M+ disabled Indians [1], [2], promoting equitable AI towards frictionless, situational communication.

\section{CONCLUSION}
This paper introduced \textit{Sanvaad}, a multimodal accessibility framework aimed at improving communication for both deaf and visually impaired users. By combining MediaPipe-based landmark extraction with an efficient residual MLP classifier, the system delivers real-time ISL recognition on edge devices without relying on heavy computational resources. Alongside this, the voice-to-sign and summarization modules provide practical speech-to-text and text-to-speech support using lightweight NLP components.

With an accuracy of 84\% on 35 gesture classes and stable performance in everyday settings, Sanvaad shows that meaningful ISL assistance can be achieved even with modest hardware. At the same time, the framework has limitations: the vocabulary of dynamic signs is still small, continuous signing remains difficult, and background noise can affect speech components.

Overall, Sanvaad represents a step toward more inclusive, practical, and locally relevant accessibility tools, and highlights how lightweight AI systems can support fair communication in real-world environments.

\end{document}